\documentclass[english,letterpaper]{article}
\usepackage[T1]{fontenc}
\usepackage[latin9]{inputenc}
\usepackage{array}
\usepackage{booktabs}
\usepackage{url}
\usepackage{amsmath}
\usepackage{amssymb}
\usepackage{graphicx}
\PassOptionsToPackage{normalem}{ulem}
\usepackage{ulem}

\makeatletter

\providecommand{\tabularnewline}{\\}

\@ifundefined{date}{}{\date{}}
\usepackage{microtype}
\usepackage{graphicx}
\usepackage{subfigure}
\usepackage{booktabs} 

\usepackage{hyperref}

\usepackage[accepted]{arxiv}

\usepackage{amsmath}
\usepackage{amssymb}
\usepackage{mathtools}
\usepackage{amsthm}

\usepackage[capitalize,noabbrev]{cleveref}

\theoremstyle{plain}

\theoremstyle{definition}

\theoremstyle{remark}

\icmltitlerunning{Tighter Bounds on the Information Bottleneck with Application to Deep Learning}

\makeatother

\usepackage{babel}
\begin{document}
\twocolumn[ \icmltitle{Tighter Bounds on the Information Bottleneck with Application to Deep Learning}
\icmlsetsymbol{equal}{*}
\begin{icmlauthorlist} \icmlauthor{Nir Weingarten}{runi} \icmlauthor{Zohar Yakhini}{runi} \icmlauthor{Moshe Butman}{runi} \icmlauthor{Ran Gilad-Bachrach}{tau} \end{icmlauthorlist}
\icmlaffiliation{runi}{Efi Arzi School of Computer Science, Reichman University, Herzliya, Israel} \icmlaffiliation{tau}{Department of Biomedical Engineering, Tel-Aviv University, Tel-Aviv, Israel}
\icmlcorrespondingauthor{Nir Weingarten}{nir.weingarten@runi.ac.il}
\vskip 0.3in ]
\printAffiliationsAndNotice{}
\begin{abstract}
Deep Neural Nets (DNNs) learn latent representations induced by their
downstream task, objective function, and other parameters. The quality
of the learned representations impacts the DNN's generalization ability
and the coherence of the emerging latent space. The Information Bottleneck
(IB) provides a hypothetically optimal framework for data modeling,
yet it is often intractable. Recent efforts combined DNNs with the
IB by applying VAE-inspired variational methods to approximate
bounds on mutual information, resulting in improved robustness to
adversarial attacks. This work introduces a new and tighter variational
bound for the IB, improving performance of previous IB-inspired DNNs.
These advancements strengthen the case for the IB and its variational
approximations as a data modeling framework, and provide a simple
method to significantly enhance the adversarial robustness of classifier
DNNs.
\end{abstract}

\section{Introduction}

In recent years, Deep Neural Networks (DNNs) have gained prominence
in various learning tasks, revolutionizing many computational fields
with their ability to approximate complex functions. Despite the great
achievements, it is still postulated that the current networks are
prone to overfit the training data \cite{Ying2019}, may be considerably
uncalibrated \cite{Guo2017} and are susceptible to adversarial attacks
\cite{Goodfellow2015}. A question emerges regarding the extraction
of an optimal representation for all data points from a restricted
set of training examples. Classic information theory provides tools
to optimize compression and transmission of data, but it does not
provide methods to gauge the relevance of a compressed signal to its
downstream task. Methods such as rate-distortion \cite{Blahut1972}
regard all information as equal, not taking into account which information
is more relevant without constructing complex distortion functions.
The Information Bottleneck (IB) \cite{Tishby1999} resolves this limitation
by defining mutual information between the learned representation
and the downstream task as a universal distortion function. Under
this definition, an optimal rate-disotrion ratio can be implicitly
computed for a Lagrange multiplier $\beta$, controlling the tradeoff
between the desired rate and distortion. However, optimizing over
the IB requires mutual information computations, which are tractable
in discrete settings and for some specific continuous distributions.
Adopting the IB framework for DNNs requires computing mutual information
for unknown distributions and has no analytic solution. However, recent
work approximated tractable upper bounds for the IB functional in
DNN settings using variational approximations. Variational Auto Encoders
(VAEs) \cite{Kingma2014} use stochastic DNNs to approximate intractable
distributions, as elaborated in Section~\ref{subsec:variational_approximations}.
Similarly to VAEs, \citet{Alemi2017} proposed using stochastic DNNs
as variational approximations of latent models, thus making possible
the computations of upper bounds for mutual information between the
DNN's input, output and latent representation. A proposed DNN optimization
method called Deep Variational Information Bottleneck (VIB) derives
an upper bound for the IB objective and minimizes its approximation
by fitting some training dataset. Optimizing classifier DNNs with
the VIB objective results in a slight decrease in test set accuracy
compared to deterministic DNNs, but yields a significant increase
in robustness to adversarial attacks.

In this study, we adopt the same information theoretic and variational
approach proposed in VIB. The work begins by deriving a new upper
bound for the IB functional. We then employ a tractable variational
approximation for this bound, named VUB - 'Variational Upper Bound'
and show that it is a tighter bound on the IB objective than VIB.
We proceed to show empirical evidence that VUB substantially increases
test set accuracy over VIB while providing similar or superior robustness
to adversarial attacks across several challenging tasks and different
modalities. Finally, we discuss these effects in the context of previous
work on the IB and on DNN regularization. The conclusion drawn is
that while increasing mutual information between encoding and output
does not necessarily improve classification, and while increasing
encoding compression does not always enhance regularization \cite{Amjad2020},
the application of IB approximations as objectives to DNNs empirically
improves regularization, suggesting better data modeling. This notion
contributes for the adaptation of the IB, and its variational approximations,
as an objective for learning tasks and as a theoretic framework to
gauge and explain data modeling.

In addition, we demonstrate that VUB can be easily adapted to any
classifier DNN, including transformer based NLP classifiers, to substantially
increase robustness to adversarial attacks while only slightly decreasing,
or in some cases even increasing, test set accuracy.

\subsection{Preliminaries}

The following literature review and derivations refer to information
theory and variational approximations. A preliminary mutual ground
and notation is provided.

We denote random variables (RVs) with upper cased letters $X,Y$,
and their realizations in lower case $x,y$. Denote discrete Probability
Mass Functions (PMFs) with an upper case $P(x)$ and continuous Probability
Density Functions (PDFs) with a lower case $p(x)$. Hat notation denotes
empirical measurements. 

Let $X,Y$ be two observed random variables with unknown distributions
$p^{*}(x),p^{*}(y)$ that we aim to model. Assume $X,Y$ are governed
by some unknown underlying process with a joint probability distribution
$p^{*}(x,y)$. We can attempt to approximate these distributions using
a model $p_{\theta}$ with parameters $\theta$ such that for generative
tasks $p_{\theta}(x)\approx p^{*}(x)$ and for discriminative tasks
$p_{\theta}(y|x)\approx p^{*}(y|x)$, using a dataset $\mathcal{S}=\left\{ (x_{1},y_{1}),...,(x_{N},y_{N})\right\} $
to fit our model. One can also assume the existence of an additional
unobserved RV $Z\sim p^{*}(z)$ that influences or generates the observed
RVs $X,Y$. Since $Z$ is unobserved it is absent from the dataset
$\mathcal{S}$ and so cannot be modeled directly. Denote $\int p_{\theta}(x|z)p_{\theta}(z)dz$
the marginal, $p_{\theta}(z)$ the prior as it is not conditioned
over any other RV, and $p_{\theta}(z|x)$ the posterior following
Bayes' rule. 

When modeling an unobserved variable of an unknown distribution we
encounter a problem as the marginal $p_{\theta}(x)=\int p_{\theta}(x,z)dz$
doesn't have an analytic solution. This intractability can be overcome
by choosing some tractable parametric variational distribution $q_{\phi}(z|x)$
to approximate the posterior $p_{\theta}(z|x)$ such that $q_{\phi}(z|x)\approx p_{\theta}(z|x)$,
and estimate $p_{\theta}(x,z)$ or $p_{\theta}(x,z|y)$ by fitting
the dataset $\mathcal{S}$ \cite{Kingma2019}.

In this work information theoretic functions share the same notation
for discrete and continuous settings. For brevity, we will only present
the continuous form:
\begin{flushleft}
\begin{tabular}{>{\raggedright}m{2.5cm}>{\raggedright}m{5cm}}
Entropy & $H_{p}(X)=-\int p(x)log\left(p(x)\right)dx$\tabularnewline
\midrule
\addlinespace[0.2cm]
Cross

Entropy & $CE(p,q)=-\int p(x)log\left(q(x)\right)dx$\tabularnewline
\midrule
\addlinespace[0.2cm]
KL

Divergence & $D_{KL}\left(p\big|\big|q\right)=\int p(x)log\left(\frac{p(x)}{q(x)}\right)dx$\tabularnewline
\midrule
\addlinespace[0.2cm]
Mutual Information & $I(X;Y)=\int\int p(x,y)log\left(\frac{p(x,y)}{p(x)p(y)}\right)dxdy$\tabularnewline
\bottomrule
\addlinespace[0.2cm]
\end{tabular}
\par\end{flushleft}

\section{Related work}

\subsection{IB and its analytic solutions}

\begin{figure}
\begin{centering}
\includegraphics[scale=0.3]{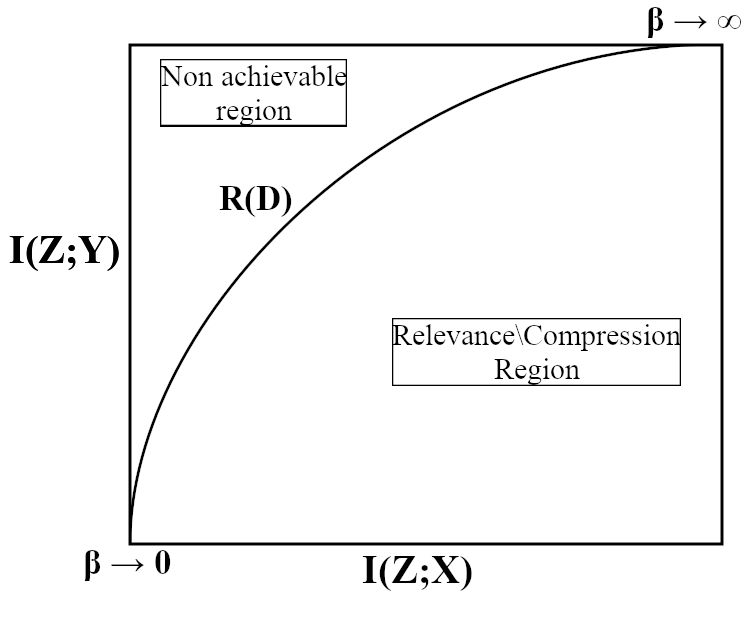}
\par\end{centering}
\caption{The information plane and curve: rate-distortion ratio over $\beta$.
At $\beta=0$ the representation is compressed but uninformative (maximal
compression), at $\beta\rightarrow\infty$ the representation is informative
but potentially overfitted (maximal information). Adapted from \cite{Slonim2002}.\label{info_plane}}
\end{figure}

Classic information theory offers rate-distortion \cite{Blahut1972}
to mitigate signal loss during compression. Rate being the signal's
compression measured by mutual information between input and output
signals, and distortion a chosen task-specific function. The Information
Bottleneck (IB) method \cite{Tishby1999} extends rate-distortion
by replacing the tailored distortion functions with mutual information
between the learned representation and the downstream task. Denote
$X$ the source signal, $Z$ its encoding and $Y$ the target signal
for some specific task. Assuming a latent variable model that follows
the Markov chain $Z\leftrightarrow X\leftrightarrow Y$, we define
some positive minimal threshold $D$ for the desired distortion. We
seek an optimal encoding $Z:\underset{P(Z|X)}{min}I(X;Z)$ subject
to $I(Z;Y)\ge D$. This constrained problem can be implicitly optimized
by minimizing the functional $L_{P(Z|X)}=I(Z;X)-\beta I(Z;Y)$, the
first term being rate and the second distortion modulated by the Lagrange
multiplier $\beta$. The optimal solution is a function of $\beta$
and was named 'the information curve' as illustrated in Figure~\ref{info_plane}.
The IB can be interpreted as a method to learn a representation that
holds just enough information to satisfy a desired task, while discarding
all other available information, presumably providing a model with
the least possible complexity.

The IB functional is only tractable when mutual information can be
computed and was originally demonstrated for soft clustering tasks
over a discrete and known distribution $P^{*}(x,y)$. \citet{Chechik2003}
extended the IB for gaussian distributions and \citet{Painsky2017}
offered a limited linear approximation of the IB for any distribution.

\subsection{IB and deep learning}

\begin{figure}
\noindent \begin{centering}
\includegraphics[scale=0.36]{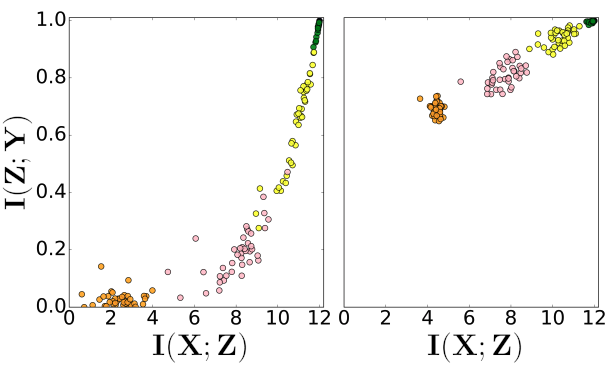}
\par\end{centering}
\caption{Information plane scatters of different DNN layers (colors) in 50
randomized networks. From \citet{ShwartzZiv2017}. Left are initial
weights, Right are at 400 epochs. Our study reproduced similar yet
repetitive behavior on complicated high dimensional tasks, as elaborated
in Section~\ref{subsec:Evaluation-and-analysis} and in Figure~\ref{estimated_info_plane}.
\label{shwartz_ziv_info_plane}}
\end{figure}

\citet{Tishby2015} proposed an IB interpretation of DNNs, regarding
them as Markov cascades of intermediate representations between hidden
layers. Under this framework, comparing the optimal and the achieved
rate-distortion ratios between DNN layers will indicate if a model
is too complex or too simple for a given task and training set. \citet{ShwartzZiv2017}
visualized and analyzed the information plane behavior of DNNs over
a toy problem with a known joint distribution. Mutual information
of the different layers was estimated and used to analyze the training
process. The learning process over Stochastic Gradient Descent (SGD)
exhibited two separate and sequential behaviors: A short Empirical
Error Minimization phase (ERM) characterized by a rapid decrease in
distortion, followed by a long compression phase with an increase
in rate until convergence to an optimal IB limit as demonstrated in
Figure~\ref{shwartz_ziv_info_plane}. Similar yet repetitive behavior
was observed in the current study, as elaborated in Section~\ref{subsec:Evaluation-and-analysis}.

\citet{Amjad2020} pointed out three flaws in the usage of the IB
functional as an objective for deterministic DNN classifiers: (1)
When data $X$ is absolutely continuous the mutual information term
$I(X;Z)$ is infinite; (2) When data $X$ is discrete the IB functional
is a piecewise constant function of the parameters, making it's SGD
optimization difficult or impossible; (3) Equivalent representations
might yield the same IB loss while one achieved better classification
rate than the other. These discrepancies were attributed to mutual
information's invariance to invertible transformations and to the
absence of a decision function in the objective.

\subsection{Variational approximations to the IB objective\label{subsec:variational_approximations}}

\citet{Kingma2014} introduced the Variational Auto Encoder (VAE)
- a stochastic generative DNN. An unobserved RV $Z$ is assumed to
generate evidence $X$ and the true probability $p^{*}(x)$ can be
modeled using a parametric model over the marginal $p_{\theta}(x)=\int p_{\theta}(x|z)p_{\theta}(z)dz$.
However, since the marginal is intractable a variational approximation
$q_{\phi}(z|x)\approx p_{\theta}(z|x)$ is proposed instead. The log
probability $log\left(p_{\theta}(x)\right)$ is then developed in
to the tractable VAE loss comprised of the Evidence Lower Bound (ELBO)
and KL regularization terms: $\underset{\text{\emph{\text{\emph{ELBO}}}}}{\mathcal{L}}=\mathbb{E}_{q_{\phi}(z|x)}\left[log\left(p_{\theta}(x|z)\right)\right]-D_{KL}\left(q_{\phi}(z|x)\big|\big|p_{\theta}(z)\right)$.
$q_{\phi}(z|x)$ is modeled using a stochastic neural encoder having
it's final activation used as parameters for the assumed variational
distribution (typically a spherical gaussian with parameters $\mu,\Sigma$).
Each forward pass emulates a stochastic realization $z\in Z$ from
these parameters by using the 'reparameterization trick': $z=\mu+\epsilon\cdot\Sigma$
for some unparameterized scalar $\epsilon\sim N(0,1)$ such that a
backwards pass is possible. \citet{Higgins2017} later proposed the
$\beta$-autoencoder, introducing a hyper parameter $\beta$ over
the KL term to control the regularization-reconstruction tradeoff.
\citet{Alemi2018} found that as the ELBO loss in VAEs depends solely
on image reconstruction it does not necessarily induce a better quality
modeling of the marginal $p_{\theta}(z)$, hence not necessarily a
better representation learned. This gap is attributed to powerful
decoders being overfitted, as will be further discussed in Section~\ref{sec:Discussion}.

\citet{Alemi2017} introduced the Variational Information Bottleneck
(VIB) as a variational approximation for an upper bound to the IB
objective for classifier DNN optimization. Bounds for $I(Z,Y)$ and
$I(X,Z)$ are derived from the non negativity of KL divergence and
are used to form an upper bound for the IB functional. This upper
bound is approximated using variational approximations for $p^{*}(y|z),\,p^{*}(z)$
as done in VAEs. This approximation of an upper bound for the IB objective
is empirically estimated as cross entropy and a beta scaled KL regularization
term as in $\beta$-autoencoders, and is optimized over the training
data using Monte Carlo sampling and the reparameterization trick.
VIB was evaluated over image classification tasks and, while causing
a slight reduction in test set accuracy, generated substantial improvements
in robustness to adversarial attacks.

Additional noteworthy contributions to this field have been made in
recent years by \citet{Achille2018,Wieczorek2019,Fischer2020} and
others. However, a detailed review of these works is beyond the scope
of this paper.

\subsection{Non IB information theoretic regularization}

Label smoothing \cite{Szegedy2016} and entropy regularization \cite{Pereyra2017}
both regularize classifier DNNs by increasing the entropy of their
output. This is done either directly by inserting a scaled conditional
entropy term to the loss function, $-\gamma\cdot H\left(p_{\theta}(y|x)\right)$,
or by smoothing the training data labels. Applying both methods was
demonstrated to improve test accuracy and model calibration on various
challenging classification tasks. In the current work a similar conditional
entropy term emerges from the derivation of the new upper bound for
the IB objective, as shown in Section~\ref{sec:vub}.

\section{From VIB to VUB\label{sec:vub}}

The VIB loss consists of a cross entropy term and a KL regularization
term, as in VAE loss. The KL term is derived from a bound on the IB
rate term $I(X;Z)$, while the cross entropy term from a bound on
the IB distortion term $I(Z;Y)=H(Y)-H(Y|Z)$. When deriving the latter
the entropy term $H(Y)$ is ignored as it is constant and does not
effect optimization. We note that since $Y$ is unknown any optimization
over $Z$, including cross entropy, depends on our decoder model of
$Y$. Following this logic, instead of omitting $H(Y)$ we replace
it with a variational approximation of the decoder entropy, which
provides a lower bound.

\subsection{IB upper bound}

We begin by establishing a new upper bound for the IB functional by
bounding the mutual information terms, using the same method as in
VIB.

Consider $I(Z;X)$:

\begin{align}
I(Z;X)= & \int\int p^{*}(x,z)log\left(p^{*}(z|x)\right)dxdz\nonumber \\
- & \int p^{*}(z)log\left(p^{*}(z)\right)dz\label{eq:i_z_x1}
\end{align}

For any probability distribution $r$ we have that $D_{KL}\left(p^{*}(z)\big|\big|r(z)\right)\ge0$,
it follows that:

\begin{equation}
\int p^{*}(z)log\left(p^{*}(z)\right)dz\ge\int p^{*}(z)log\left(r(z)\right)dz\label{eq:d_kl}
\end{equation}

And so by Equation~\ref{eq:d_kl}:

\begin{equation}
I(Z;X)\le\int\int p^{*}(x)p^{*}(z|x)log\left(\frac{p^{*}(z|x)}{r(z)}\right)dxdz\label{eq:i_z_x2}
\end{equation}

Consider $I(Z;Y)$:

For any probability distribution $c$ we have that $D_{KL}\left(p^{*}(y|z)\big|\big|c(y|z)\right)\ge0$,
it follows that:

\begin{equation}
\int p^{*}(y|z)log\left(p^{*}(y|z)\right)dy\ge\int p^{*}(y|z)log\left(c(y|z)\right)dy\label{eq:d_kl2}
\end{equation}

And so by Equation~\ref{eq:d_kl2}:

\begin{align}
I(Z;Y)= & \int\int p^{*}(y,z)log\left(\frac{p^{*}(y,z)}{p^{*}(y)p^{*}(z)}\right)dydz\nonumber \\
\ge & \int\int p^{*}(y|z)p^{*}(z)log\left(\frac{c(y|z)}{p^{*}(y)}\right)dydz\nonumber \\
= & \int\int p^{*}(y,z)log\left(c(y|z)\right)dydz+H_{p^{*}}(Y)\label{eq:i_z_y}
\end{align}

We now diverge from the original VIB derivation by replacing $H_{p^{*}}(Y)$
with $H_{c}(Y|Z)$ instead of omitting it. In addition, we limit the
new term to make sure that the inequality $H(Y|Z)\le H(Y)$ holds
when computing entropy over the different distributions $p^{*}$ and
$c$.

\begin{align}
I(Z;Y)\ge & \int\int p^{*}(y,z)log\left(c(y|z)\right)dydz\nonumber \\
+ & min\left\{ H_{p^{*}}(Y),H_{c}(Y|Z)\right\} \label{eq:i_z_y2}
\end{align}

We further develop this term using the IB Markov chain $Z\leftrightarrow X\leftrightarrow Y$
and total probability:

\begin{align}
 & I(Z;Y)\ge\nonumber \\
 & \int\int\int p^{*}(x)p^{*}(y|x)p^{*}(z|x)log\left(c(y|z)\right)dxdydz\nonumber \\
- & min\left\{ H_{p^{*}}(Y),-\int\int c(y,z)log\left(c(y|z)\right)dydz\right\} \label{eq:i_z_y3}
\end{align}

Finally, we define a new upper bound for the IB functional named $L_{UB}$
by joining the bound on rate in Equation \ref{eq:i_z_x2} with the
bound on distortion in Equation \ref{eq:i_z_y3}:

\begin{align}
 & L_{UB}\equiv\nonumber \\
 & \int\int p^{*}(x)p^{*}(z|x)log\left(\frac{p^{*}(z|x)}{r(z)}\right)dxdz\nonumber \\
- & \int\int\int p^{*}(x)p^{*}(y|x)p^{*}(z|x)log\left(c(y|z)\right)dxdydz\nonumber \\
+ & min\left\{ H_{p^{*}}(Y),-\int\int c(y,z)log\left(c(y|z)\right)dydz\right\} \label{eq:l_ub}
\end{align}

It is easy to verify that the bound holds for all $\beta\ge0$ such
that $L_{IB}=\beta\cdot I\left(Z;X\right)-I\left(Z;Y\right)$.

\subsection{Variational approximation}

Following the same variational approach as in VIB, we define $L_{VUB}$
as a new tractable upper bound for the IB functional. Let $p^{*}(x,y,z)$
be the unknown joint distribution, $e(z|x)$ a variational encoder
approximating $p^{*}(z|x)$ and $c(y|z)$ a variational classifier
approximating $p^{*}(y|z)$:

\begin{align}
 & L_{VUB}\equiv\nonumber \\
 & \beta\int\int p^{*}(x)e(z|x)log\left(\frac{e(z|x)}{r(z)}\right)dxdz\nonumber \\
- & \int\int\int p^{*}(x)p^{*}(y|x)e(z|x)log\left(c(y|z)\right)dxdydz\label{eq:l_vub}\\
- & min\Bigg\{ H_{p^{*}}(Y),\nonumber \\
 & -\int\int\int p^{*}(x)e(z|x)c(y|z)log\left(c(y|z)\right)dxdydz\Bigg\}\nonumber \\
\ge & L_{IB}\nonumber 
\end{align}

\subsection{Empirical estimation}

We proceed to model VUB using DNNs and optimize it using Monte Carlo
sampling over some training dataset. Let $e_{\phi}$ be a stochastic
DNN encoder with parameters $\phi$ applying the reparameterization
trick such that $e_{\phi}(x)\sim N(\mu,\Sigma)$ and let $C_{\lambda}$
be a discrete classifier DNN parameterized by $\lambda$ such that
$C_{\lambda}(\hat{z})\sim Categorical$.

\begin{align}
\hat{L}_{VUB} & \equiv\nonumber \\
 & \frac{1}{N}\sum_{n=1}^{N}\Bigg[\beta\cdot D_{KL}\left(e_{\phi}(x_{n})\big|\big|r(z)\right)\label{eq:l_vub_empirical}\\
- & P^{*}(y_{n})\cdot log\left(C_{\lambda}\left(e_{\phi}(x_{n})\right)\right)\nonumber \\
- & min\left\{ H(\hat{Y}),H\left(C_{\lambda}\left(e_{\phi}(x_{n})\right)\right)\right\} \Bigg]\nonumber 
\end{align}

As in VIB and VAE, $e_{\phi}(x)$ and $r(z)$ are computed as spherical
gaussians. $e_{\phi}(x)$ by using the first half of the encoder's
output entries as $\mu$ and the second as the diagonal $\Sigma$,
and $r(z)$ by a standard normal gaussian.

\subsection{Interpretation}

Similarly to the confidence penalty suggested by \citet{Pereyra2017},
the new derivation adds classifier regularization to the VIB objective.
Regularizing the classifier might prevent it from overfitting, and
is a possible remedy to the discrepancies in the ELBO loss observed
by \citet{Alemi2018}, as elaborated in Section~\ref{sec:Discussion}.

In terms of tightness we have that VUB is a tighter theoretical bound
on the IB objective than VIB for any $Y$ such that $H(Y)>0$, and
a tighter empirical bound for all $Y$.

\section{Experiments\label{sec:Experiments}}

We follow the experimental setup proposed by \citet{Alemi2017}, extending
it to NLP tasks as well. Image classification models were trained
on the ImageNet 2012 dataset \cite{deng2009imagenet} and text classification
over\textbf{ }the IMDB\textbf{ }sentiment analysis dataset \cite{maas2011imdb}.
For each dataset, a competitive pre-trained model (Vanilla model)
was evaluated and then used to encode embeddings. These embeddings
were then used as a dataset for a new stochastic classifier net with
either a VIB or a VUB loss function. Stochastic classifiers consisted
of two ReLU activated linear layers of the same dimensions as the
pre-trained model's logits (2048 for image and 768 for text classification),
followed by reparameterization and a final softmax activated FC layer.
Learning rate was $10^{-4}$ and decaying exponentially with a factor
of 0.97 every two epochs. Batch sizes were 32 for ImageNet and 16
for IMDB. We used a single forward pass per sample for inference.
Each model was trained and evaluated 5 times per $\beta$ value with
consistent performance. Beta values of $\beta=10^{-i}$ for $i\in\{1,2,3\}$
were tested since previous studies indicated this is the best range
for VIB \cite{Alemi2017,Alemi2018}. Each model was evaluated using
test set accuracy and robustness to various adversarial attacks over
the test set. For image classification we employed the untargeted
Fast Gradient Sign (FGS) attack \cite{Goodfellow2015} as well as
the targeted CW $L_{2}$ optimization attack \cite{Carlini2017}, \cite{Kaiwen2018}.
For text classification we used the untargeted Deep Word Bug attack
\cite{gao2018deepwordbug} as well as the untargeted PWWS attack \cite{ren2019pwws}, \cite{Morris2020}.
All models were trained using an Nvidia RTX3080 GPU. Code to reconstruct
the experiments is available in the following github repository: \url{https://github.com/hopl1t/vub}.

\subsection{Image classification}

A pre-trained inceptionV3 \cite{Szegedy2016} base model was used
and achieved a 77.21\% accuracy on the ImageNet 2012 validation set
(Test set for ImageNet is unavailable). Note that inceptionV3 yields
a slightly worse single shot accuracy than inceptionV2 (80.4\%) when
run in a single model and single crop setting, however we've used
InceptionV3 over V2 for simplicity. Each model was trained for 100
epochs. The entire validation set was used to measure accuracy and
robustness to FGS attacks, while only 1\% of it was used for CW attacks
as they are computationally expensive.

\subsubsection{Evaluation and analysis}

\begin{figure}[h]
\begin{centering}
\includegraphics[scale=0.38]{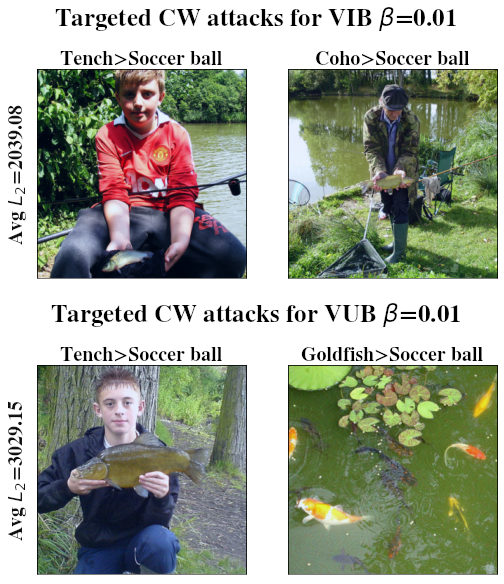}
\par\end{centering}
\caption{Successful targeted CW attack examples. Images are perturbations of
previously successfully classified instances from the ImageNet validation
set. The target label is 'Soccer ball'. Average $L_{2}$ distance
required for a successful attack is shown on the left. The higher
the required $L_{2}$ distance the greater the visible change required
to fool the model. Original and wrongly assigned labels are listed
at the top of each image. Mind the difference in noticeable change
as compared to the FGS perturbations presented in Figure~\ref{fig:untargeted_examples},
and between VIB and VUB perturbations.\label{fig:targeted_examples}}
\end{figure}

\begin{figure}[h]
\begin{centering}
\includegraphics[scale=0.25]{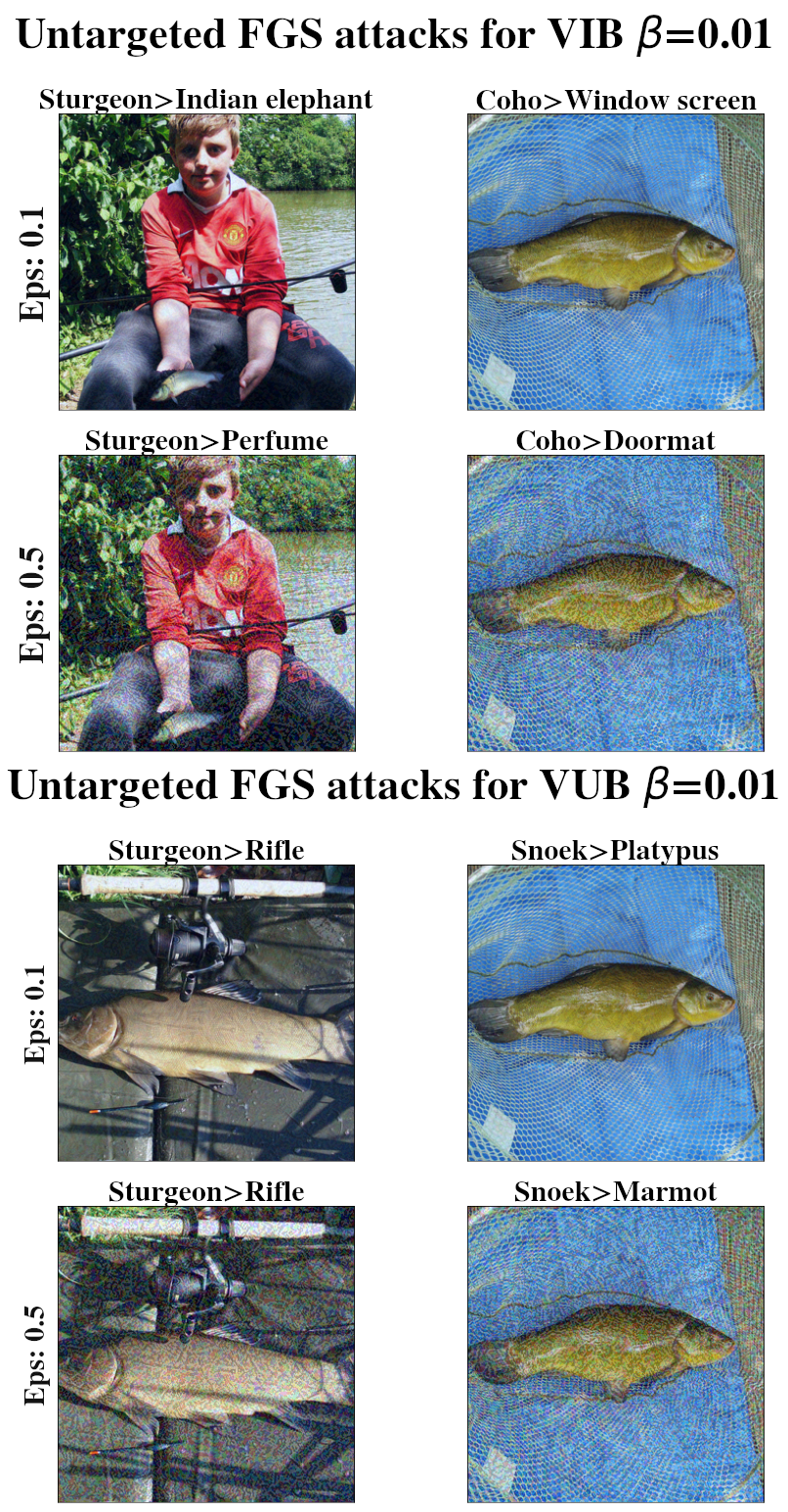}
\par\end{centering}
\caption{Successful untargeted FGS attack examples. Images are perturbations
of previously successfully classified instances from the ImageNet
validation set. Perturbation magnitude is determined by the parameter
$\epsilon$ shown on the left, the higher the more perturbed. Original
and wrongly assigned labels are listed at the top of each image. Notice
the deterioration of image quality as $\epsilon$ increases. \label{fig:untargeted_examples}}
\end{figure}

\begin{table}
\begin{centering}
\begin{tabular*}{8cm}{@{\extracolsep{\fill}}>{\centering}m{0.027\paperheight}>{\centering}m{0.027\paperheight}>{\centering}m{0.027\paperheight}>{\centering}m{0.027\paperheight}>{\centering}m{0.027\paperheight}}
\toprule 
\textbf{$\beta$} & \textbf{Val $\uparrow$} & \textbf{$\underset{\epsilon=0.1}{\text{FGS}}\downarrow$} & \textbf{$\underset{\epsilon=0.5}{\text{FGS}}\downarrow$} & \textbf{CW$\uparrow$}\tabularnewline
\midrule
\midrule 
\multicolumn{5}{c}{\textbf{Vanilla model}}\tabularnewline
\midrule 
- & \textbf{77.2\%} & \textbf{68.9\%} & \textbf{67.7\%} & \textbf{788}\tabularnewline
\midrule 
\multicolumn{5}{c}{\textbf{VIB models}}\tabularnewline
\midrule 
$10^{-3}$ & \textbf{73.7\%}

$\pm.1\%$ & \textbf{59.5\%}

$\pm.2\%$ & \textbf{63.9\%}

$\pm.2\%$ & \textbf{3917}

$\pm291$\tabularnewline
\midrule 
$10^{-2}$ & \textbf{72.8\%}

$\pm.1\%$ & \textbf{53.5\%}

$\pm.2\%$ & \textbf{62.0\%}

$\pm.1\%$ & \textbf{3318}

$\pm293$\tabularnewline
\midrule 
$10^{-1}$ & \textbf{72.1\%}

$\pm.01\%$ & \textbf{58.4\%}

$\pm.1\%$ & \textbf{62.0\%}

$\pm.1\%$ & \textbf{3318}

$\pm293$\tabularnewline
\midrule 
\multicolumn{5}{c}{\textbf{VUB models}}\tabularnewline
\midrule 
$10^{-3}$ & \textbf{75.5\%}

$\pm.03\%$ & \textbf{62.8\%}

$\pm.1\%$ & \textbf{66.4\%}

$\pm.1\%$ & \textbf{2666}

$\pm140$\tabularnewline
\midrule 
$10^{-2}$ & \textbf{75.0\%}

$\pm.05\%$ & \textbf{57.6\%}

$\pm.2\%$ & \textbf{64.3\%}

$\pm.1\%$ & \textbf{1564}

$\pm218$\tabularnewline
\midrule 
$10^{-1}$ & \textbf{74.8\%}

$\pm0.09\%$ & \textbf{57.9\%}

$\pm.5\%$ & \textbf{64.8\%}

$\pm.5\%$ & \textbf{3575}

$\pm456$\tabularnewline
\bottomrule
\end{tabular*}
\par\end{centering}
\caption{ImageNet evaluation scores for vanilla, VIB and VUB models, average
over 5 runs with standard deviation. First column is performance on
the ImageNet validation set (higher is better $\uparrow$), second
and third columns are the \% of successful FGS attacks at $\epsilon=0.1,0.5$
(lower is better $\downarrow$) and the fourth column is the average
$L_{2}$ distance for a successful Carlini Wagner $L_{2}$ targeted
attack (higher is better $\uparrow$).\label{tab:imagenet_evaluation}}
\end{table}
Image classification evaluation results are shown in Table~\ref{tab:imagenet_evaluation},
examples of successful attacks are shown in Figures~\ref{fig:targeted_examples}, \ref{fig:untargeted_examples}.
The empirical results presented in Table \ref{tab:imagenet_evaluation}
confirm that while VIB and VUB reduce performance on the validation
set, they substantially improve robustness to adversarial attacks.
Moreover, these results demonstrate that VUB significantly outperforms
VIB in terms of validation set accuracy while providing competitive
robustness to attacks. A comparison of the best VIB and VUB models
further substantiates these findings, with statistical significance
confirmed by a p-value of less than 0.05 in a Wilcoxon rank sum test.

\begin{figure}
\begin{centering}
\includegraphics[scale=0.31]{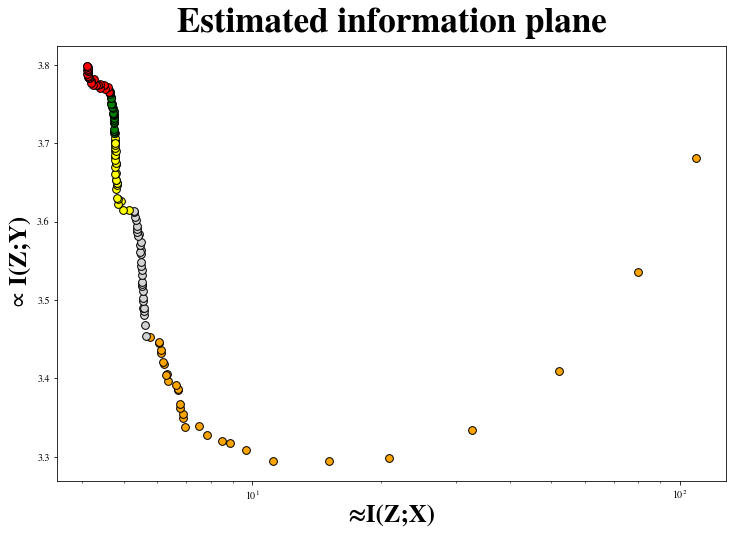}
\par\end{centering}
\caption{Estimated information plane metrics per epoch for VUB trained on the
IMDB dataset with $\beta=0.001$. $I(Z;X)$ is approximated by $H(R)-H(Z|X)$
and $\frac{1}{CE(Y;\hat{Y})}$ is used as an analog for $I(Z;Y)$.
The epochs have been grouped and color-coded in intervals of 30 epochs
in the order: Orange (0-30), gray (30-60), yellow (60-90), green (90-120)
and red (120-150). We notice recurring patterns of distortion reduction
followed by rate increase, resembling the ERM and representation compression
stages described by \citet{ShwartzZiv2017}. \label{estimated_info_plane}}
\end{figure}

\subsection{Text classification}

A fine tuned BERT uncased \cite{Devlin2019} base model was used and
achieved a 93.0\% accuracy on the\textbf{ }IMDB sentiment analysis
test set. Each model was trained for 150 epochs. The entire test set
was used to measure accuracy, while only the first 200 entries in
the test set were used for adversarial attacks as they are computationally
expensive.

\subsubsection{Evaluation and analysis\label{subsec:Evaluation-and-analysis}}

Text classification evaluation results are shown in Table~\ref{tab:text_attack_evaluation},
examples of successful attacks are shown in Figure~\ref{tab:text_attack_examples}.
In this modality VUB significantly outperforms VIB in both test set
accuracy and robustness to both attacks. Moreover, VUB also outperomed
the original model in terms of test set accuracy. A comparison of
the best VIB and VUB models further substantiates these findings,
with statistical significance confirmed by a p-value of less than
0.05 in a Wilcoxon rank sum test.

In addition to the above evaluation metrics, we also measured approximated
rate and distortion throughout training and plotted them on the information
curve as shown in Figure~\ref{estimated_info_plane}. We notice recurring
patterns of distortion reduction followed by rate increase, resembling
the ERM and representation compression stages described by \citet{ShwartzZiv2017}.\textbf{ }

\begin{table}
\begin{centering}
\begin{tabular*}{8cm}{@{\extracolsep{\fill}}>{\centering}p{0.04\paperheight}>{\centering}p{0.04\paperheight}>{\centering}p{0.04\paperheight}>{\centering}p{0.04\paperheight}}
\toprule 
\textbf{$\beta$} & \textbf{Test$\uparrow$} & \textbf{DWB$\downarrow$} & \textbf{PWWS$\downarrow$}\tabularnewline
\midrule
\midrule 
\multicolumn{3}{c}{\textbf{Vanilla model}} & \tabularnewline
\midrule 
- & \textbf{93.0\%} & \textbf{54.3\%} & \textbf{100\%}\tabularnewline
\midrule 
\multicolumn{3}{c}{\textbf{VIB models}} & \tabularnewline
\midrule 
$10^{-3}$ & \textbf{91.0\%}

$\pm1.0\%$ & \textbf{35.1\%}

$\pm4.4\%$ & \textbf{41.6\%}

$\pm6.6\%$\tabularnewline
\midrule 
$10^{-2}$ & \textbf{90.8\%}

$\pm0.5\%$ & \textbf{41.0\%}

$\pm4.8\%$ & \textbf{62.9\%}

$\pm14.3\%$\tabularnewline
\midrule 
$10^{-1}$ & \textbf{89.4\%}

$\pm.9\%$ & \textbf{90.0\%}

$\pm8.0\%$ & \textbf{99.1\%}

$\pm0.9\%$\tabularnewline
\midrule 
\multicolumn{3}{c}{\textbf{VUB models}} & \tabularnewline
\midrule 
$10^{-3}$ & \textbf{93.2\%}

$\pm.5\%$ & \textbf{27.5\%}

$\pm2.0\%$ & \textbf{28.4\%}

$\pm1.3\%$\tabularnewline
\midrule 
$10^{-2}$ & \textbf{92.6\%}

$\pm.8\%$ & \textbf{30.8\%}

$\pm2.0\%$ & \textbf{50.0\%}

$\pm4.8\%$\tabularnewline
\midrule 
$10^{-1}$ & \textbf{89.2\%}

$\pm2.0\%$ & \textbf{99.2\%}

$\pm0.5\%$ & \textbf{100\%}

$\pm0\%$\tabularnewline
\bottomrule
\end{tabular*}
\par\end{centering}
\caption{Evaluation for vanilla, VIB and VUB models, average over 5 runs with
standard deviation over the IMDB dataset. First column is performance
on the test set (higher is better $\uparrow$), second is \% of successful
Deep Word Bug attacks (lower is better $\downarrow$) and the third
column is \% of successful PWWS attacks (lower is better $\downarrow$).
\label{tab:text_attack_evaluation}}
\end{table}
\begin{table}
\begin{centering}
\begin{tabular}{>{\centering}p{8cm}}
\toprule 
\textbf{Text perturbed with DWB}\tabularnewline
\midrule
\midrule 
g\textbf{\uline{n}}reat historical movie, will not allow a viewer
to leave once you begin to watch. View is presented differently than
displayed by most school books on this s\textbf{\uline{S}}ubject
{[}...{]}\tabularnewline
\midrule 
\textbf{Text perturbed with PWWS}\tabularnewline
\midrule 
the acting , costumes , music , cinematography and sound are all \sout{astounding}\textbf{\uline{dumbfounding}}
given the production 's austere locales .\tabularnewline
\bottomrule
\end{tabular}
\par\end{centering}
\caption{Examples of successful DWB and PWWS perturbations on a vanilla Bert
model fine tuned over the IMDB dataset. The original input strings
were perturbed such that inserted tokens are marked in underscored
boldface and removed tokens in strikethrough. Both examples were classified
correctly as 'Positive sentiment' before the attack and 'Negative
sentiment' afterwards. \label{tab:text_attack_examples}}
\end{table}

\section{Discussion\label{sec:Discussion}}

While providing a complete framework for optimal data modeling, the
IB, and it\textquoteright s variational approximations, rely on three
assumptions: (1) It suffices to optimize the mutual information metric
to optimize a model\textquoteright s performance; (2) Forgetting more
information about the input while keeping the same information about
the output induces better generalization; (3) Mutual information between
the input, output and latent representation can be either computed
or approximated to a desired level of accuracy. Our study strengthens
the argument for using the Information Bottleneck combined with variational
approximations to obtain robust models that can withstand adversarial
attacks. By deriving a tighter bound on the IB functional, we demonstrate
it's utility as the Variational Upper Bound (VUB) objective for neural
networks. We demonstrate that VUB outperforms the Variational Information
Bottleneck (VIB) in terms of test accuracy while providing similar
or superior robustness to adversarial attacks in challenging classification
tasks of different modalities, suggesting an improvement in data modeling
quality.

Comparing VIB and VUB we observe that both methods promote a disentangled
latent space by using a stochastic factorized prior, as suggested
by \citet{Chen2018}. In addition, both methods utilize KL regularization,
enforcing clustering around a $0$ mean which might increase latent
smoothness. These traits can make it difficult for minor perturbations
to significantly alter latent semantics, making the models more robust
to attacks. In the case of VUB, the enhanced results induced by classifier
regularization not only reinforce previous studies on the ELBO function,
which suggest that overly powerful decoders diminish the quality of
learned representations \cite{Alemi2018}, but also align with the
confidence penalty proposed by \citet{Pereyra2017}. 

In addition, we observed that in many cases VIB achieves lower validation
set cross entropy while VUB achieves significantly higher test set
accuracy. We attribute this gap to the VUB models becoming more calibrated,
and we suggest that practitioners also monitor validation set accuracy
and rate-distortion ratio during training. These metrics may be more
informative indicators of model performance than validation set cross
entropy alone, as validation cross entropy could increase as models
become more calibrated. 

We made another interesting observation during our study regarding
information plane behavior throughout the training process. While
previous research has documented the occurrence of error minimization
and representation compression phases, our work revealed that these
phases can occur in cycles throughout training. This finding is particularly
noteworthy because previous studies observed this phenomenon in simple
toy problems, whereas our research demonstrated it in complex tasks
of high dimensionality with unknown distributions. This suggests that
this information plane behavior is not limited to simplified scenarios
but is a characteristic of the learning process in more challenging
tasks as well. 

In conclusion, while the IB and its variational approximations do
not provide a complete theoretical framework for DNN data modeling
and regularization, they offer a strong, measurable and theoretically
grounded approach. VUB is presented as a tractable and tighter upper
bound of the IB functional that can be easily adapted to any classifier
DNN, including transformer based text classifiers, to significantly
increase robustness to various adversarial attacks while inflicting
minimal decrease in test set performance, and in some cases even increasing
it. 

This study opens many opportunities for further research. Besides
further improvements to the upper bound, it is intriguing to use VUB
in self-supervised learning and in generative tasks. Other possible
directions, including measuring model calibration as proposed by \citet{Achille2018}
are left for future work.

\clearpage{}

\section*{Impact}

This paper presents work whose goal is to advance the field of Machine
Learning. There are many potential societal consequences of our work,
none which we feel must be specifically highlighted here.

\bibliography{bib.bib}
\end{document}